\documentclass{article}

\usepackage{microtype}
\usepackage{graphicx}
\usepackage{subfigure}
\usepackage{booktabs} 

\usepackage{hyperref}

\usepackage[accepted]{icml2023}

\usepackage{amsmath}
\usepackage{amssymb}
\usepackage{mathtools}
\usepackage{amsthm}

\usepackage[capitalize,noabbrev]{cleveref}

\theoremstyle{plain}

\theoremstyle{definition}

\theoremstyle{remark}

\usepackage[textsize=tiny]{todonotes}

\graphicspath{{figures/}}

\icmltitlerunning{Large Language Model Programs}

\begin{document}

\twocolumn[
\icmltitle{Large Language Model Programs}



\icmlsetsymbol{equal}{*}

\begin{icmlauthorlist}
\icmlauthor{Imanol Schlag}{idsia}
\icmlauthor{Sainbayar Sukhbaatar}{meta}
\icmlauthor{Asli Celikyilmaz}{meta}
\icmlauthor{Wen-tau Yih}{meta}
\icmlauthor{Jason Weston}{meta}
\icmlauthor{Jürgen Schmidhuber}{idsia,kaust}
\icmlauthor{Xian Li}{meta}
\end{icmlauthorlist}

\icmlaffiliation{idsia}{The Swiss AI Lab IDSIA, SUPSI \& USI}
\icmlaffiliation{kaust}{King Abdullah University of Science and Technology}
\icmlaffiliation{meta}{Meta AI}

\icmlcorrespondingauthor{Imanol Schlag}{imanol@idsia.ch}

\icmlkeywords{Machine Learning, ICML}

\vskip 0.3in
]



\printAffiliationsAndNotice{}  

\begin{abstract}
In recent years, large pre-trained language models (LLMs) have demonstrated the ability to follow instructions and perform novel tasks from a few examples.
The possibility to parameterise an LLM through such in-context examples widens their capability at a much lower cost than finetuning. 
We extend this line of reasoning and present a method which further expands the capabilities of an LLM by embedding it within an algorithm or program.
To demonstrate the benefits of this approach, we present an illustrative example of evidence-supported question-answering.
We obtain a 6.4\% improvement over the chain of thought baseline through a more algorithmic approach without any finetuning.
Furthermore, we highlight recent work from this perspective and discuss the advantages and disadvantages in comparison to the standard approaches.
\end{abstract}

\section{Introduction}
\label{sec_introduction}

Scaling language models to hundreds of billions of parameters (LLMs) and training them on terabytes of text data has led to state-of-the-art performance on a large variety of natural language processing tasks. Additionally, it has also led to the emergent ability to learn a new skill merely from instructions or a few examples, as seen in GPT-3 \cite{gpt3}.
Despite this, LLMs struggle to display algorithmic abilities, such as sorting or searching, even when finetuned on traces of such \cite{anil2022exploring, valmeekam2022large, liu2023transformers, deletang2023neural}.

Finetuning a model on human traces (i.e. imitation learning) is a common strategy to infuse a model with complex behaviours.
For example, in the context of LLMs, recent work trains on expert demonstrations of interacting with a browser \cite{nakano2021webgpt} or a webshop \cite{yao2022webshop} in order to improve their respective tasks.
However, there are reasons to question the correctness of the algorithms learned from such examples.
First, the decoder-only Transformer language model \cite{vaswani2017attention} is not computationally universal because it can only condition on a finite length input \cite{fan2021addressing}.
Second, previous work demonstrated that the decoder-only architecture struggles to learn even simple programs through gradient descent that would generalise out of distribution \cite{anil2022exploring, deletang2023neural}. 

Besides finetuning, in-context learning may be leveraged to improve a model's ability to execute algorithms. 
E.g., recently \citet{zhou2023teaching} introduce a prompt construction which improves the LLM's ability to perform arithmetic algorithms. 
A key characteristic of such approaches is the use of a single call to the model and the use of a single prompt. 
As a consequence, the method is limited by the size of the input and the necessity to prompt engineer multiple algorithm steps within one prompt.
The authors report that this can lead to interference between steps which hurts the final performance. 

As an alternative, we propose embedding LLMs into a program or algorithm.
Crucially, instead of the LLM being responsible for maintaining the current state of the program (i.e. its context), the LLM, for each step of the program, is only presented with a step-specific prompt and context.
Hiding information which is irrelevant to the current step allows us to focus on isolated subproblems whose results are further combined in future calls to the LLM. 
This intuitive approach allows us to extend the ability of an LLM to more complex tasks which are currently too difficult either because of a lack of ability or an architectural constraint such as an insufficiently large context.
Concurrent work proves that embedding an LLM within a recurrent algorithm which enables the interaction with an external memory component makes the system computationally universal \cite{schuurmans2023memory}.

\begin{figure*}[ht]
\begin{center}
\centerline{\includegraphics[width=0.8\textwidth]{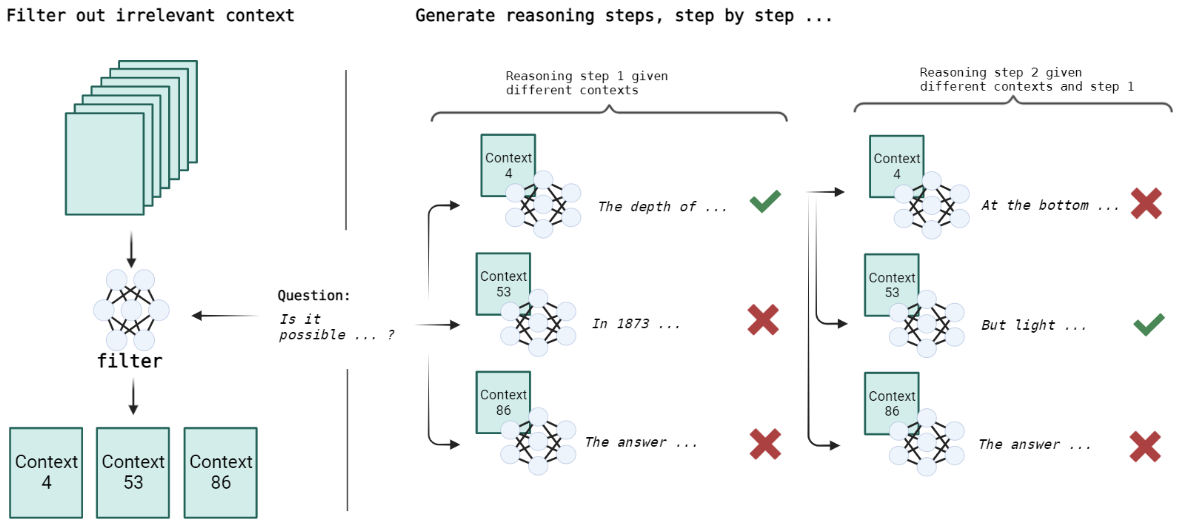}}
\caption{An illustration of our two-part LLM program example. The first part (left) of our program filters out irrelevant paragraphs from a large set. This is achieved using an LLM and the current question. The second part (right) depicts the generation of the individual reasoning steps until they result in an answer. The tree search is over the choice of paragraph used within the context when generating a single step in the reasoning chain. We expand the reasoning chain which ranks the highest as described in Section \ref{sec_search}.}
\label{fig_illustration}
\end{center}
\vskip -0.5cm
\end{figure*}

Both approaches, embedding and finetuning, incorporate domain-specific information and thus lead to a more specialised model.
Embedding the model in an algorithm, or \textit{program with an LLM}, forces the LLM to follow a high-level procedure (whatever it may be) which may limit its flexibility and requires a high-level understanding of the correct procedure for a given task. 
On the other hand, finetuning requires the recognition, and possibly generation, of domain-specific training data which requires a similar level of understanding. 
We argue that there are situations where it may be more promising to embed the LLM in a program explicitly instead of training on traces of a desired program.
Due to the difficulty and cost of training LLMs we have already seen problem-specific LLM programs with no or very targeted training lead to better performance on specific problems such as reasoning \cite{kazemi2022lambada}, question-answering \cite{zhou2023leasttomost}, text-summarization \cite{Wu2021Recursively}, text-generation \cite{yang2022re3}, among others \cite{zeng2023socratic, Xu2021BeyondGM, karpas2022mrkl}.

In this work, we present the advantages and disadvantages of programming with LLMs and present a general approach which we call a \textit{Large Language Model Program} (Section \ref{sec_method}).
In Section \ref{sec_problem}, we then present an LLM program example for evidence-supported question answering.
In order to accurately answer a question, the program first extracts the important facts from the knowledge sources and seamlessly incorporates them into the reasoning of its response.
A simple illustration of the program is given in Figure \ref{fig_illustration}.
We then go on to describe how our general approach can be applied to various other settings, described in Section \ref{sec_further}.

\section{Limitations of LLMs and the Benefit of Programming With Them}
\label{sec_method}
LLMs are a product of three necessary ingredients: massive parallel compute, large amounts of data, and a highly parallelisable language modelling architecture. To this date, the architecture deployed by all LLMs of 175B parameters or more is the decoder-only Transformer \cite{vaswani2017attention}. The ability to parallelise allows for efficient training of large models and the use of large amounts of data prevents it from overfitting. In fact, when training LLMs they rarely see any training data twice. 

LLMs which were trained on internet-scale data are impressive text generators that can produce plausible paragraphs in the form of short stories, reviews, summaries, poems, etc. Such \textit{raw} LLMs are trained on data that includes high-quality texts such as those from Wikipedia or published books but also subjectively low-quality texts with toxic and biased content which then leads to LLMs that are to a certain degree toxic and biased themselves \cite{dev2020measuring, sheng2021societal}. 

Another drawback of training on random internet text is that such raw LLMs also struggle to be conversational, follow instructions, use tools, or interact with an environment. Although finetuning on examples of such behaviour has been shown to improve the model's ability \cite{ouyang2022training, wei2022finetuned, gupta2022improving}, such an approach is difficult to scale due to the lack of such data and may not generalise systematically due to the possibility of leveraging algorithmic short-cuts 
 \cite{liu2023transformers}. 

Another limitation of LLMs arises from their decoder-only Transformer architecture, which has a finite context that restricts their capability to only process information within the predefined context.
While this architecture may be advantageous for tasks like translation or language modelling, it struggles to learn certain classes of algorithms accurately, which can affect its ability to generalize beyond its training distribution \cite{csordas2021devil}.

Such issues are likely going to be present also at a larger scale if an LLM is, either implicitly or explicitly, finetuned on traces of algorithms.
One example is presented by \citet{anil2022exploring} who finetune LLMs of up to 64B parameters on a variation of parity and report large drops in performance for just slightly longer (or shorter) sequences than it was trained on. 
It is worthwhile to mention that it has long been known that recurrent neural networks, such as the Long Short-Term Memory (LSTM), can generalise perfectly on parity and other counter-based context-sensitive languages \cite{hochreiter1997long, gers2001lstm, deletang2023neural}.

To overcome such limitations, we propose \textit{LLM programs}, a general approach to enhance the capability of an LLM-based system. 
Using an LLM program, we recognise the limitation of the LLM as a general agent. 
Instead of further training the model, we recursively deconstruct the expected behaviour into simpler steps that the LLM can perform to a sufficient degree. 
These individual steps are then strung together by a classic computer program (such as Python) that parses the outputs of previous steps, uses control flow, and augments the prompts of succeeding steps. 

One important distinction of LLM programs from previous work that relies on external systems like calculators \cite{thoppilan2022lamda} is that the current state of the program is not maintained by the LLM but by the program in which the LLM is embedded. 
Consequently, every call to the LLM only includes the information that is necessary for the particular step.
There are several advantages to LLM programs:
\begin{itemize}
    \setlength\itemsep{0.05cm}
    \item Embedding an LLM in a program can significantly expand the theoretical and practical capabilities of the system with no or little finetuning and can help the system generalise more systematically.
    \item LLM programs incorporate high-level algorithmic information by breaking down a complex task into multiple simpler steps with little or no training. In contrast, learning to perform a complex task from examples alone may require a large amount of compute and high-quality data.
    \item Understanding the problem decomposition allows for more fine-grained input and output specifications for each subproblem or step of the program. These specifications allow for a more manageable and systematic evaluation of the LLM's performance by testing the model on each subproblem in isolation, either through a test set or manual inspection.
    \item An LLM program performs many queries to the model instead of just one. Each query may contain a step-specific prompt that improves the model's performance for that particular step. Prompt engineering in a setting with a single query to the LLM may find it much more challenging to provide an effective prompt because examples or descriptions for different steps may interfere with each other and the prompt overall will take up more space within its already limited context.
\end{itemize}

In summary, LLM programs offer a promising way to address the limitations of LLMs and expand their practical utility in various domains. In the next section, we will take the reader through an example of an LLM program for the purpose of question-answering.

\section{An LLM Program for Question Answering}
\label{sec_problem}

\paragraph{Motivating example.}
Imagine using an LLM as an interface for a company's website. 
The LLM was not trained on that website, which means that it would not be able to answer questions about the company's newest products, such as its features, specifications, or availability, or about the opening hours of a new store.
In this example of reading comprehension, an LLM would perform poorly because it has not been trained on the company's website nor can it fit the entire website into its context.
Finetuning the LLM may not be practical, as training these models is costly and websites often undergo changes (e.g. a product may unexpectedly become unavailable) which would require the model to be constantly retrained.
Additionally, it is unclear to which extent the finetuned model would be faithful to the facts of the finetuning data.

A more practical approach is to use a frozen LLM that has exclusive access to the latest version of the website where each page represents one document.
Since we cannot load all documents at once, we may construct a program to carefully select the documents necessary to answer a user's question.


\paragraph{The dataset.}

We construct a dataset to develop an LLM program which can answer questions by leveraging many additional knowledge sources. 
To this end, we take advantage of the StrategyQA dataset \cite{Geva2021DidAU}. 
The StrategyQA dataset is a binary classification benchmark for implicit multi-step question-answering.
The reasoning traces in StrategyQA are more difficult than previous question-answering datasets due to different forms of question decomposition and the need for detailed knowledge from different domains.
Consider e.g. the question \textit{Can sunlight travel to the deepest part of the Black Sea?}. 
In order to answer this question the LLM needs to know that the deepest part of the Black Sea is about 2k meters whereas sunlight only penetrates water up to about 1k meters.
Having that information, inferring the answer becomes significantly easier for an LLM .
See our experiments in Table \ref{tbl_answer_results} where the LLM performs best when it has all the necessary facts in its context.

We build a question-answering dataset from StrategyQA where every question is supported with a large number of knowledge sources of which only a few are actually relevant to the current question.
Most of them will be randomly sampled from other questions and are thus very unlikely to contain any useful information.
Crucially, even just a handful of those paragraphs would be too long for the context length of current LLMs.

The StrategyQA dataset in total consists of 2780 questions annotated with their decomposition and per-step evidence paragraphs sourced from Wikipedia. 
About 918 of the questions in StrategyQA come with evidence paragraphs that support all reasoning steps.
We will limit ourselves to this subset to ensure that the additional knowledge sources actually contain all information necessary to infer the answer.
Otherwise, it would require the LLM to have learned the necessary facts from its pre-training data, which we do not assume to be the case.

Previous work reported mixed performances for question-answering on the regular StrategyQA benchmark \cite{chowdhery2022palm}.
This might be because the facts used for individual reasoning steps are very rare and unlikely to be remembered or because the reasoning is too complex. 
While improved reasoning methods (such as chain of thought \cite{wei2022chain, nye2022show, kojima2022large, wang2023selfconsistency}) have shown little performance gain on this dataset with models of 175B parameters \cite{srivastava2022beyond, taylor2022galactica}, larger models and models which have been trained longer do seem to improve \cite{chowdhery2022palm, hoffmann2022an}. 
This may indicate that the lack of knowledge is more problematic on this dataset than the ability to reason.

\paragraph{The LLM.}
In our experiments, we use OPT-175B \cite{zhang2022opt}, a raw language model of 175B parameters trained on 300B tokens from a large mix of news articles, books, Pushshift.io Reddit previously compiled by a third party, and the Pile \cite{gao2020pile}.
Because it has not been finetuned on high-quality instructional or conversational data, we classify it as a raw language model.
As reported by \citet{zhang2022opt}, we find OPT-175B struggles to follow instructions. 
Nevertheless, we demonstrate that we can achieve more complex behaviour by embedding OPT within a program. 

\subsection{Developing the LLM Program}
For the problem of evidence-supported question-answering, we decompose the problem into a filtering part and a tree search part.
Given the question, the filtering part will loop over the provided paragraphs and select the most relevant ones.
In the second part, we will search over reasoning chains by generating one reasoning step at a time. 
When generating a reasoning step we can choose which one of the filtered paragraphs we want to condition on. 
Similar to a tree search, we rank the reasoning chains according to our ranking metric and expand the highest-ranked one by generating $n$ possible continuations given $n$ evidence paragraphs.
This process repeats until it produced an answer or until it reached the maximum number of steps at which point it will force the LLM to answer the question by evaluating the negative log-likelihood (NLL) of a fixed yes or no answer. 

\paragraph{An example trace.}
When answering a question like \textit{Could the members of The Police perform lawful arrests?} we use the LLM to select a few paragraphs from a much bigger collection of which many are not relevant to this particular question.
For this example, selecting the top 10 paragraphs according to OPT is sufficient to capture all the relevant information to successfully reason about the answer.
To find a strong reasoning sequence, we generate each step using a different paragraph as context. 
The tree search is over the choice of paragraph to be used as context for a particular step. 
For the question about the band \textit{The Police} the highest ranked step has a paragraph in its context which is about the members of the band. 
Given that context, the model generates the first step \textit{The members of The Police are Sting, Stewart Copeland, Andy Summers, and Henry Padovani}. 
Following the first step, the highest-ranked step is conditioned on another paragraph with more details about the band The Police in general. This results in the second step which says \textit{The members of The Police are not police officers.}
After that step, the model's highest-ranked next step is conditioned on a paragraph which explains that only police offers are allowed to arrest people. 
From that the generated next step says \textit{Thus, the members of The Police could not perform lawful arrests.}. 
Finally, the model generates the last step \textit{Thus, the answer is no.}, which is recognised as an answer and evaluated to be correct.

In the following two subsections, we go into more detail about each part of our LLM program.

\subsubsection{Evidence Filtering Experiments}
\label{sec_filter}

This section describes our approaches to the paragraph filtering part of our program and their performance. Developing and evaluating each call to the LLM in isolation allows us to experiment, improve, and analyse the overall performance in a systematic way. Our first attempt will rely on few-shot examples and prompt-engineering the model to classify one paragraph given the question.

\paragraph{Blackbox prompting.}
Our first approach centres on few-shot prompting, using OPT to output a yes or no answer when asked whether a certain paragraph is relevant to a specific question.
This binary classification dataset comprises 300 samples randomly selected from the StrategyQA data.
Each sample includes a question and an associated paragraph, and the objective is to classify whether the paragraph is relevant to the question.

We carefully designed the dataset for this subproblem so that half of the questions are paired with paragraphs randomly selected from a list of evidence paragraphs, while the other half are paired with unrelated paragraphs from other questions.
This ensures an equal distribution of relevant and irrelevant paragraphs which puts the random baseline at 50\%.
Our experiments in Table \ref{tbl_filter_results_classify} indicate that neither OPT nor InstructGPT \cite{ouyang2022training} achieves satisfactory performance in classifying a single paragraph. 
Even \textit{Tk-Instruct} \cite{wang2022super}, a model which has been trained on a large variety of distinct and expert-written tasks, fails to reliably recognise relevant paragraphs. 

This is surprising since the classification task seems simple for humans: it is often inconceivable how a randomly sampled paragraph could be relevant to the question. 
Furthermore, evidence paragraphs often share words or phrases or are semantically relevant to the question.
Because of the outcome of our isolated experiment, we decide to explore an alternative approach.

\begin{table}[h]
\caption{Top-1 binary classification accuracies for a single paragraph using prompt engineering techniques. Random guessing is at 50\%. Note that using this method to classify from a large set of paragraphs will significantly reduce the accuracy.}
\label{tbl_filter_results_classify}
\begin{center}
\begin{small}
\begin{sc}
\begin{tabular}{lc}
\toprule
Method & Accuracy \\
\midrule
OPT-175B few-shot prompt                & 53.33\% \\
text-davinci-002 (InstructGPT)          & 55.67\% \\
OPT-175B few-shot + chain of thought    & 56.00\% \\
tk-instruct 11B                         & 61.60\% \\
\bottomrule
\end{tabular}
\end{sc}
\end{small}
\end{center}
\vskip -0.25cm
\end{table}

\paragraph{Likelihood evaluation.} Our second approach does not treat the LLM as a black box anymore but instead directly uses the average negative log-likelihood (NLL) of the question when following each paragraph to create a ranking of all paragraphs. To evaluate this subproblem in isolation, we constructed a new dataset. Each question now consists of 100 paragraphs where one is sampled from the list of evidence paragraphs of that question and the other 99 are sampled from other questions. In this setting, random guessing only achieves 1\% accuracy.
Notice that 100 paragraphs clearly exceed the context length of OPT.
We present the pseudo-code in Algorithm \ref{alg_ranking} (see Appendix).

\begin{figure}[t]
\begin{center}
\centerline{\includegraphics[width=\columnwidth]{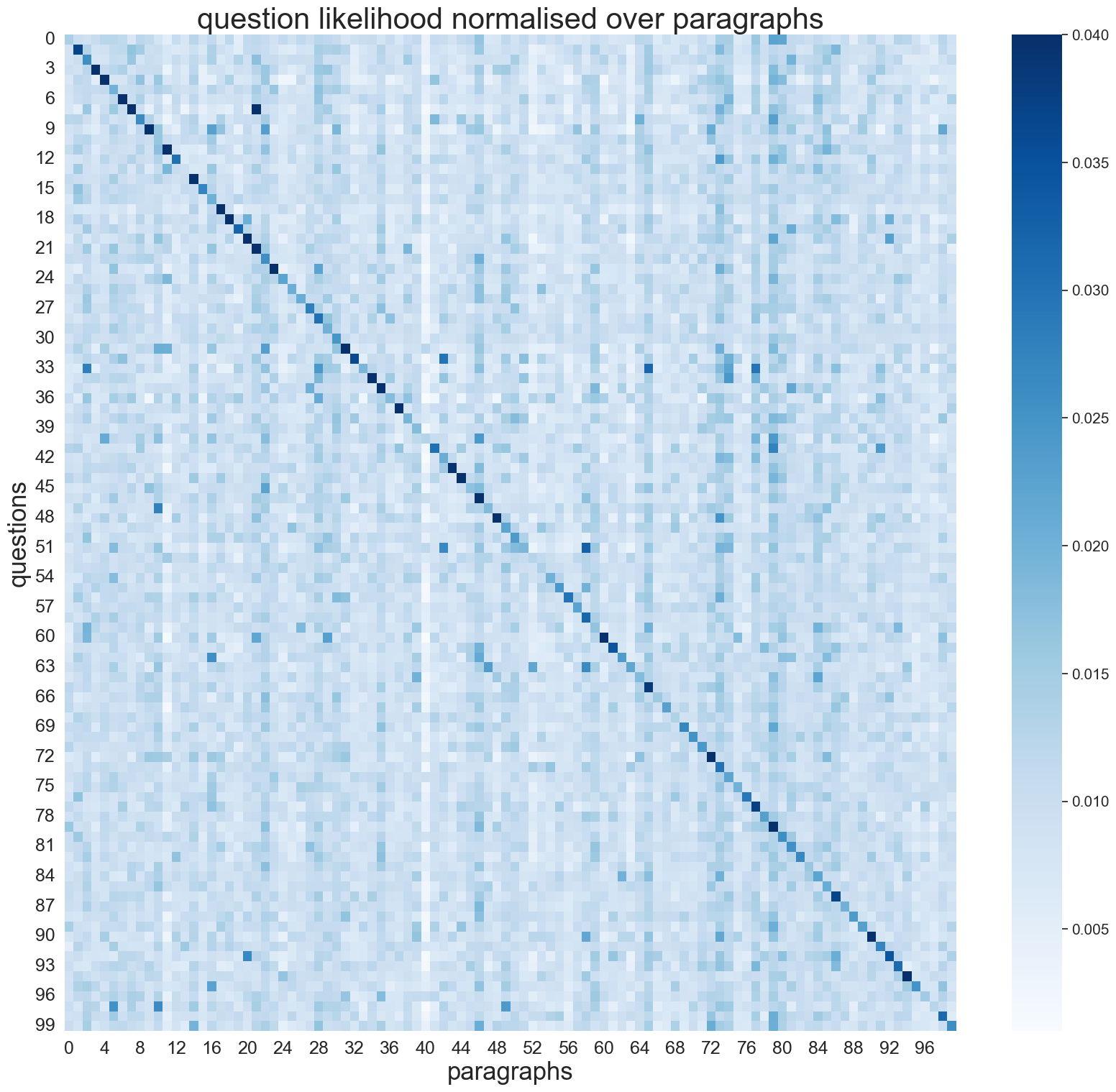}}
\caption{Average likelihood of each question (row) given a paragraph (column) normalised over paragraphs.}
\label{fig_likelihood}
\end{center}
\vskip -0.5cm
\end{figure}

\begin{figure}[t]
\begin{center}
\centerline{\includegraphics[width=\columnwidth]{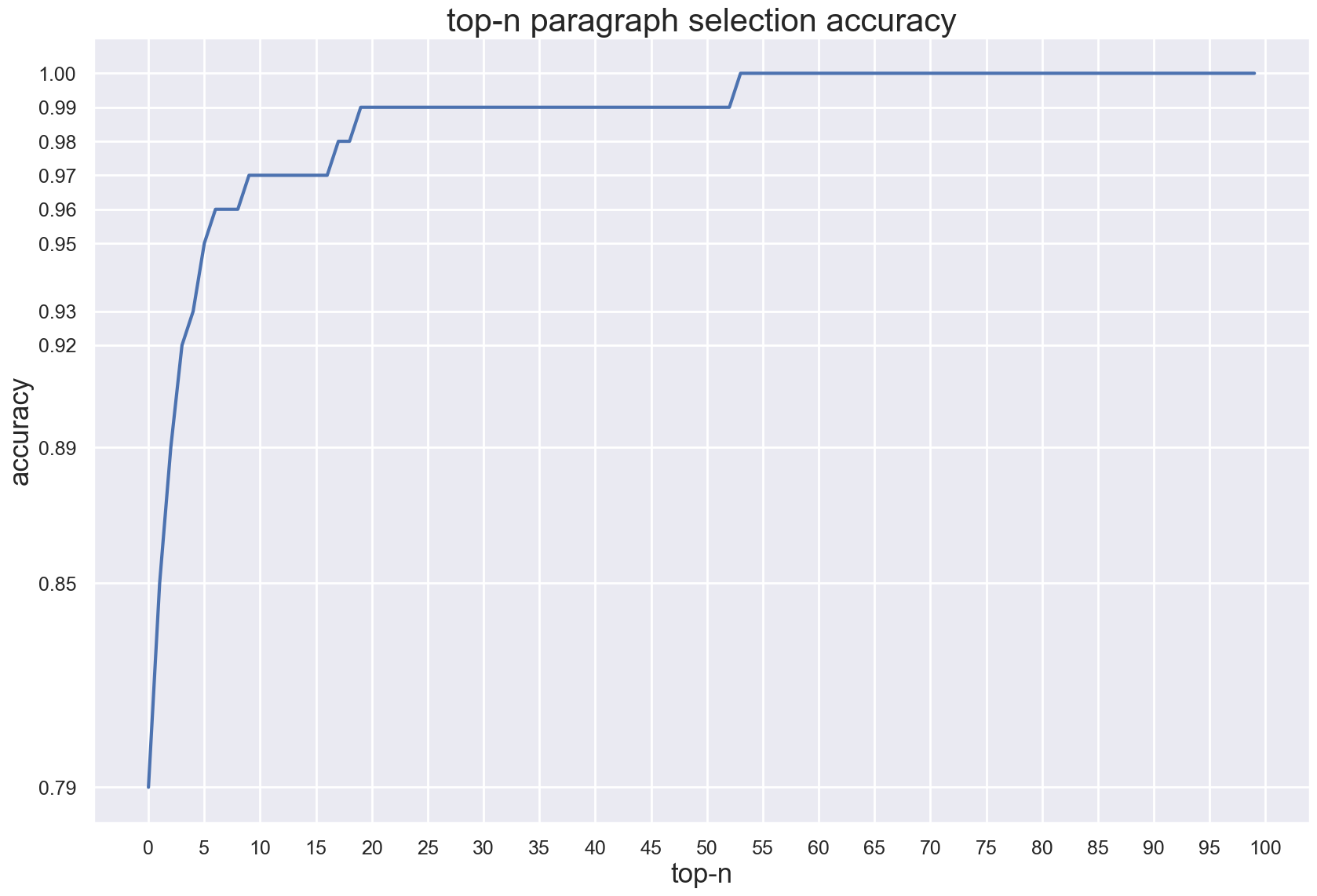}}
\caption{Top-n accuracy of selecting the true evidence paragraph from our likelihood ranking of 100 paragraphs. The random guessing baseline for top-1 is 1\%. The ranking approach significantly outperforms the in-context approach from Table \ref{tbl_filter_results_classify}.}
\label{fig_topn}
\end{center}
\vskip -0.5cm
\end{figure}

We achieve good results without a description or few-shot examples. 
We plot accuracy over top-n in Figure \ref{fig_topn} and the average likelihood of each paragraph given each question in Figure \ref{fig_likelihood}.
Top-1 accuracy is $0.79$ and top-5 is already at $0.93$ which shows that the ranking approach is vastly more effective than the blackbox prompting approach despite the simplicity of the task and the instruction-finetuning of some of the models.
It also serves as an example of the advantage of embedding an LLM within a program because the LLM does not have access to the NLL of its own outputs and, thus, this superior solution would have not been possible otherwise.

\subsubsection{Tree Search Experiment}
\label{sec_search}
In the previous subsection, we presented the filtering part of our program: measuring the relevancy of each paragraph, ordering them, and selecting the top $n$. 
It trivially generalises in a systematic way to more paragraphs since the processing of the list of paragraphs is done by a Python for-loop and doesn’t solely rely on an LLM's ability to implement it reliably.
This is especially powerful if we know what the optimal algorithm or behaviour should be and if we want to have control over it.

In this section, we present a more systematic search over reasoning steps to ensure generalisation where we consider it most crucial. 
As in the filtering part, we develop and test the tree search in isolation using a dedicated dataset. 

Raw LLMs do not have access to an external system when deducing answers to difficult questions.
This means that the model can only rely on the noisy and inevitably incomplete and outdated knowledge that has been stored in its weights. The questions in the StrategyQA dataset seem to often require knowledge that is not present in LLMs. 
For this reason, we introduce an evidence-supported chain of thought where the reasoning chain is generated over multiple steps and each step has a different support paragraph within its context. 
This allows the model to use information from the paragraph to generate reasoning steps it would otherwise be unable to produce. 
For StrategyQA most questions rely on reasoning steps which draw from different knowledge domains. 
In order to find the most valuable paragraph for each reasoning step, we perform a tree search where the root node is the question and each level of the tree spans over the set of possible evidence paragraphs used as context to generate the next step. 
Every path through the tree is a (possibly incomplete) reasoning chain. 
Searching over all possible reasoning chains is infeasible which is why we rank all reasoning chains and continually expand the highest-ranked chain. 
We present pseudo-code in Algorithm \ref{alg_treesearch} (see Appendix).

\newcommand{\PDelta}{\ensuremath{\Delta_\text{P}~}}
\newcommand{\QDelta}{\ensuremath{\Delta_\text{Q}~}}

We explore two ranking strategies. The first is the average NLL of the chain so far (where the average NLL of each step $S$ is computed when conditioning on its respective paragraph). With this approach, the model will expand the reasoning chain which has been the most likely so far. This approach works reasonably well but can lead to issues. We find that if an LLM directly copies entire phrases from the evidence paragraph $P$, they will be given a very low NLL. This can lead to repetitions or ongoing deductive steps which quickly end up in arguing about things that are irrelevant to the question $Q$. 
Our second ranking strategy is an approach which tries to mitigate such issues by ranking the generated reasoning steps by their average NLL \textit{difference}: with and without the paragraph (\PDelta), and with and without the question (\QDelta). 
Those length-normalised differences, \PDelta and \QDelta, allow us to select reasoning chains which leverage the provided context (which reduces hallucinations) but remain on-topic (which reduces divergent reasoning). 
\begin{align}
    \text{nll}(x) &= -\log(x) / \text{len}(x) \\
    \PDelta &= \text{nll}(p(S|Q,P)) - \text{nll}(p(S|Q)) \\
    \QDelta &= \text{nll}(p(S|Q,P)) - \text{nll}(p(S|P))
\end{align}
%
%
%
%
Generated reasoning steps with a negative \PDelta rely more strongly on the paragraph and are thus more likely to incorporate information from it.
Generated reasoning steps with a negative \QDelta rely more strongly on the question which leads to less divergence and favours steps which remain on topic.  
We find that steps conditioned on paragraphs where $\PDelta + \QDelta$ is the lowest leads to better reasoning chains and improves accuracy. 

\begin{table*}[h]
\caption{Binary classification accuracy on the StrategyQA subset with fully supported evidence. CoT stands for chain of thought. Golden facts are the facts necessary to answer the question according to the StrategyQA dataset.}
\label{tbl_answer_results}
\begin{center}
\begin{small}
\begin{sc}
\begin{tabular}{lc}
\toprule
Method & Accuracy \\
\midrule
OPT-175B, few-shot, no-CoT                            & 50.33\% \\
OPT-175B, few-shot, with-CoT                          & 60.11\% \\
OPT-175B, few-shot, with-tree-search, NLL ranking     & 65.98\% \\
OPT-175B, few-shot, with-tree-search, Delta ranking   & 66.41\% \\
\midrule
OPT-175B, few-shot, with-golden-facts                 & 81.27\% \\
OPT-175B, few-shot, with-golden-facts, with-CoT       & 81.12\% \\
\bottomrule
\end{tabular}
\end{sc}
\end{small}
\end{center}
\vskip -0.25cm
\end{table*}

To evaluate the tree search in isolation, we assume that the evidence paragraphs contain information about all reasoning steps necessary to answer the question. This is not the case for all StrategyQA questions which is why we limit ourselves to the 918 StrategyQA questions which are fully supported.

Note that this technique is a variation of beam search \cite{medress1977speech}, i.e. a heuristic search algorithm which explores a graph by expanding the highest-ranking node in a limited set. 
It is quite different from the usual use of beam search for generating text with language models where in order to generate the next token we select the $k$ tokens with the highest probability given the same context. 
Instead, we always generate the most likely tokens but search over the limited set of $k$ paragraphs to be used as context.

Our experimental results in Table \ref{tbl_answer_results} demonstrate that the model’s reasoning ability has improved over the OPT chain of thought baseline.
To better highlight the advantage of our tree search we include a baseline where we add the golden facts to the context of OPT using a custom prompt.
Golden facts are statements provided by the authors of StrategyQA which should contain the factual knowledge necessary to deduce the correct answer.
With the golden facts in its context, OPT achieves about 81.2\%. 
This can be seen as a performance upper bound for the OPT model since it represents the setting where the model has all the necessary facts clearly within its context. 

\section{Further LLM Program Examples}
\label{sec_further}
Training an LLM is a costly and complex engineering challenge.
For this reason, training and research on such models is mostly done in well-funded industrial labs.
However, in recent years, GPT-3 \cite{gpt3} has become easily accessible to the public through its API \cite{brockman2018openaiapi} and the weights of LLMs from the OPT \cite{zhang2022opt} and BLOOM \cite{scao2022bloom} family have been made publicly available, sparking a flurry of LLM research.

Some recent and concurrent work implicitly follows the presented LLM Program approach. 
Such work often has the characteristic that it describes stages or an algorithm and parameterises different steps with different prompts.
In this Section, we'll go through all recent and concurrent work that uses this emerging methodology.


\citet{creswell2022faithful} decompose the inference of an answer into a recurrent algorithm which consists of a selection and inference step, as well as a halting criterion. 
Each step then consists of a finetuned LLM which only has access to the information necessary in order to prevent the model from learning shortcuts, such as directly predicting the answer from the question instead of performing a deductive step.
The final system's capability of finding proofs is increased by over 20\% in comparison with baselines. 

\citet{yang2022language} present a modular approach to extract rules from natural language facts. 
To achieve this they use an LLM to generate many candidate rules based on input facts and the desired rule template. 
In four further steps, the same LLM with a different prompt is used to filter out rules which do not meet the necessary requirements for induction (e.g. triviality).
The authors present experimental results for the entire system as well as each module in isolation which indicate the benefits of their 5-step program.

In a similar vein, various works propose the use of \textit{verifiers} to improve the quality of the generated samples through some form of repeated verification \cite{cobbe2021training}.
For example, \citet{saparov2023language} leverages diverse prompts and verifies each individual step in a recurrent fashion, resulting in significant gains in accuracy without any finetuning.

\citet{kim2021lightbulb} present a 3-step program to identify unverifiable presuppositions in the question-answering setting.
E.g., the question \textit{Which linguist invented the lightbulb?} contains the false presupposition that a linguist has invented the lightbulb. 
The authors report that false presuppositions can make up a substantial share of the unanswerable questions in the popular Natural Questions \cite{kwiatkowski2019naturalquestions} dataset.
Given a question, their program first generates presuppositions, verifies them, and finally, generates an explanation.
The authors test different strategies where they use neural and rule-based models for different steps, testing each in isolation.

Current approaches to reasoning with language models are forward chaining, i.e. they start of with certain facts and search in the space of rules until they find a goal statement. 
This also includes our own method presented in Section \ref{sec_search}, but \citet{kazemi2022lambada} argue that backwards chaining, i.e. decomposing the goal statement until the subgoals can be proven from the facts, is heavily favoured in the classic automated proof-finding literature.
For this reason, they implement a backwards-chaining using four modules: fact check, rule selection, goal decomposition, and sign agreement.
Each of the modules is implemented using a pre-trained LLM with a custom prompt and the modules are subroutines of the \textit{LAMBADA} program. 
Thanks to the ability of the LLM, the modules remain relatively high-level which results in a relatively simple program with two nested for-loops. 
The LAMBADA approach significantly outperforms other methods such as chain of thought \cite{wei2022chain} and selection-inference \cite{creswell2022faithful}. 

Another related line of works first decomposes the original question into simpler subquestions which should be easier to answer.
With the answers to the subquestions, a final module answers the original question. 
\citet{patel2022question} use questions decomposed by humans which significantly improves performance.
\citet{zhou2023leasttomost} use an LLM to automatically decompose questions. 
Their empirical results indicate that they significantly outperform chain of thought prompting. 
However, different from previous approaches, the authors perform a question decomposition within a single call to an LLM instead of implementing a recurrent algorithm. As a result, their approaches may be more prone to mistakes on more complex examples which are out of distribution since the output specification of such a compact step includes many more possibilities than in a more broken down program like LAMBADA. For further decomposition examples see e.g. \citet{perez2020unsupervised, yang2022seqzero}.

Regular LLMs are isolated systems that can only access the information that has been transferred into their weights.
As we argued in Section \ref{sec_method}, such systems may lack the necessary information to perform certain tasks.
Leveraging an additional, possibly non-neural, system could be a viable alternative to an increase in model scale. 
E.g., \citet{lazaridou2023internetaugmented} generate a google search query whose result will be added as context to the question-answering prompt.
The improvements in factuality and accuracy indicate the benefits of "inference-type" interventions, i.e. embedding the model within a simple program which during inference allows the system to leverage results from a classic document retrieval system without the need for a dedicated retrieval-augmented model. 

Interestingly, an LLM is not always able to use the knowledge that it has stored in its weights. 
This is shown by \citet{liu2022generated} who demonstrate a simple 2-step program which first generates a question-specific set of facts before answering the question can result in better performance (for similar methods see \citet{paranjape2021prompting, li2022self}).
\citet{liu2022rainier} further improve such knowledge extraction using reinforcement learning based on increased question-answering performance.
See also the work of \citet{cohen2023crawling} on extracting a knowledge base from a language model using a multi-step program.

Apart from question-answering, we also see the emergents of LLM programs for generative tasks. 
\citet{yang2022re3} recursively prompt an LLM with the story plan and the current story state to generate coherent stories of up to 2500 words.
Similarly, \citet{Wu2021recursivelySummarizing} summarise entire books by recursively summarising previous summaries of fixed-size chunks.
Both are great examples of recursive programs which overcome the finite-context limitation of the decoder-only Transformer.

In another line of work, LLMs are embedded in an ongoing model-environment loop to plan and suggest actions to a robot given high-level goals \cite{ichter2022saycan, huang2022inner}. 
However, current approaches give all context to the LLM and do not have a mechanism to update an external memory.
As a result, the input and output specification of the model is likely to exceed the current capabilities of LLMs for certain extreme examples.

A rapidly growing body of work further generalises the notion of the environment which the LLM interacts with.
These approaches combine the strength of an LLM with the strength of another connectionist or classic system. 
Examples of such include the LLMs which repeatedly generate short executable programs as intermediate reasoning steps \cite{gao2022pal},
which generate chess commentaries with the help of symbolic reasoning engines \cite{lee2022improving}, or which use a variety of other tools such as a calculator \cite{karpas2022mrkl}. 

In the future, we will likely see more elaborate algorithms being developed which embed several neural and symbolic modules to solve increasingly complex tasks.
To support such projects, new libraries have been recently created. 
LangChain \cite{chase2023langchain} is a library which supports developers to combine LLMs with other sources of computation. GPT-Index \cite{liu2022gptindex} is a collection of data structures designed to make it easier to use external knowledge bases with LLMs.
\citet{reppert2023iterated} present ICE, an open-source tool for visualizing the execution traces of LM
programs.

Recent work also studied the benefits of LLM programs. 
\citet{Wu2022PromptChainerCL} study the user needs when authoring their own LLM programs and present PromptChainer to support building prototypes for applications. 
\citet{wu2022ai} present an interactive "chain authoring system" and study how users modify them to improve performance but also user satisfaction.

\paragraph{General purpose programs.}
Another interesting line of research is going to centre around task-\textit{independent} LLM programs which, so far, have received much less attention.
\citet{shuster2022language} present a system which augments a conversational agent with a knowledge retrieval step before responding to the user. 
Similarly, \citet{shuster2022blenderbot} present BlenderBot 3, a conversational bot based on an LLM but which is augmented with a classic long-term storage such that it can remember information about the user across several sessions (which e.g. ChatGPT \cite{schulman2022chatgpt} cannot). 
Another approach is presented by \citet{dalvi2022towards}, who use a simple program in form of a dialogue tree to learn facts from the user.
Facts are stored in a dynamic memory which the LLM also updates. 
Crucially, the program includes an interaction loop with the user which allows the user to correct the model which then updates its dynamic memory with new or corrected facts. 
Hence, this LLM program may be considered an example of an LLM program which implements a crude learning algorithm. 
This opens up an exciting direction for future research which can benefit from symbolic learning algorithms such as e.g. the Optimal Ordered Problem Solver or the G\"odel Machine \cite{schmidhuber2004optimal, schmidhuber2007godel}.

\section{Discussion and Related Work}
\label{sec_relatedwork}
The presented method of programming with pre-trained models stands in opposition to the common deep-learning philosophy of training a single omniferous black-box model which only has to be scaled in terms of parameter count and training data. 
Although central to the current success of LLMs, such an approach comes with several drawbacks: 

\begin{itemize} 
    \item The computation of deep connectionist models, such as a large pre-trained Transformer language model, is notoriously difficult to interpret (although there is ongoing research, see e.g. \citet{elhage2021mathematical}). Breaking a problem down into multiple steps can not just improve performance but also increase the interpretability of its inference.
    
    \item LLMs are trained on large amounts of text which can contain toxic, biased, or otherwise undesired content. As a result, an LLM may also output undesired text. The language model itself does not provide any safety mechanisms to prevent such outputs. Embedding an LLM within a program is a simple and effective way to include a safety mechanism which e.g. filters out unwanted LLM responses.
    
    \item Tuning an LLM on expert trajectories of complex behaviour requires large amounts of high-quality and behaviour-specific data which is difficult and expensive to acquire. Breaking the problem into subproblems may allow the identification of specific lower-level capabilities that are missing. Focusing the data collection on such \textit{blindspots} is potentially a faster and more efficient approach and lower-level capabilities may be useful for a wide range of problems. 
    
    \item It is difficult to give any guarantees for neural models due to the lack of interpretability. Furthermore, it is well-known that neural networks struggle to generalise out of distribution. However, embedding one or several connectionist modules with a program allows us to give some trivial generalisation guarantees that do not hold otherwise. E.g., the filtering and search aspect of our program, as presented in Section \ref{sec_method}, generalises trivially to a larger number of paragraphs.

    \item So far, all LLMs of 100B parameters or more are variations of the Transformer architecture. They thus inherit its limitations, such as a finite context of a few thousand tokens \cite{hutchins2022blockrecurrent}. Embedding an LLM within a task-independent program that is responsible to select and load relevant documents into context or which summarises past text may serve as a way of overcoming such limitations.
\end{itemize}

We believe that the method of programming with LLMs as presented in this work can mitigate many drawbacks in settings where the expected processing of a query is well understood.
Many recent papers have demonstrated the benefits of such an approach (see the previous Section \ref{sec_further}), but significantly fewer have taken a higher-level view. Here we mention some recent or concurrent works which are related to our general perspective.

Concurrent to our work, \citet{khot2023decomposed} present \textit{Decomposed Prompting}, a framework for decomposing tasks into subtasks which can be solved in isolation by LLMs using a subtask-specific prompt. 
The authors demonstrate better out-of-distribution performance on simple string manipulation and question-answering tasks. 

In earlier work, \citet{dohan2022language} have presented a related perspective which describes compositions of pre-trained models as probabilistic programs.
In so-called \textit{language model cascades}, LLMs are random variables of type string within a probabilistic program. 
With such a perspective the authors present a unifying view of existing algorithms such as chain of thought \cite{wei2022chain}, scratchpads \cite{nye2022show}, or STaR \cite{zelikman2022star}, which complements our more algorithmic perspective. 

Similarly, \citet{creswell2022faithful} briefly describe their approach as \textit{algorithmic prompting} where the response of a language model given the first prompt is integrated into future prompts. The authors argue that such prompt manipulations and constructions can be composed into entire algorithms to achieve more sophisticated behaviour.
In follow-up work, \citet{shanahan2022talking} argues that such an approach is necessary to build a trustworthy reasoning system.

\citet{zeng2023socratic} propose a modular framework with multiple pre-trained models which are composed to capture new multimodal capabilities without the need for end-to-end training or finetuning. 

A different but related approach describes a looped transformer model as \textit{programmable computers} \cite{giannou2023looped}. Through a simple loop they iteratively call a transformer model with a scratchpad, memory, and instructions, analogous to computer programs.

Our approach is inspired by the “learning to think” report \cite{schmidhuber2015learning} (Sec. 5.3) where a controller network $C$ learns to send sequences of activations into another network $M$ after which $C$ reads $M$'s activations in order to exploit its knowledge. Our program, however, is not a trained neural network but an explicit algorithm.

\section{Conclusion}
We have presented LLM programs, the emerging methodology of embedding pre-trained connectionist models, such as large language models, in a classic program to carry out more complex tasks.
It is central to this method to decompose the main problem recursively into subproblems until they can be solved by a single query to the model. 
With more fine-grained input and output specifications for each subproblem, the model's capabilities can be developed and tested in isolation.
We describe an example of this method in the setting of evidence-supported question-answering and demonstrate an improvement in performance without any finetuning.
We also list the advantages and disadvantages and highlight recent works from this perspective.

\bibliography{paper}
\bibliographystyle{icml2023}


\newpage
\quad
\newpage
\appendix
\section{Algorithms}
\subsection{Ranking Algorithm}
\begin{algorithm}[h]
   \caption{Ranking paragraphs by the average negative log-likelihood of the question and returning the \text{top-n}.}
   \label{alg_ranking}
\begin{algorithmic}
   \STATE {\bfseries Input:} question $q$, paragraphs $P$, $n$
   \STATE {\bfseries initialise:} nlls $= [\quad]$
   \FOR{paragraph $p$ {\bfseries in} $P$}
   \STATE $\text{avg\_nll} = \text{LLM}(q, p)$ 
   \STATE \text{nlls.append((i, \text{avg\_nll}))}
   \ENDFOR
   \STATE \textbf{return} \text{sorted(nlls)[:$n$]}
\end{algorithmic}
\end{algorithm}

\subsection{Tree Search Algorithm}
\begin{algorithm}[h]
   \caption{Tree search over steps generated conditioned on different evidence paragraphs ranked by a ranking criteria $r$.}
   \label{alg_treesearch}
\begin{algorithmic}
   \STATE {\bfseries Input:} question $q$, paragraphs $P$, criteria $r$ 
   \STATE {\bfseries initialise:} chains $= [[q]]$, complete $= [\quad]$
   \WHILE{$\text{len}(\text{complete}) <= 3$}
   \STATE $c = \text{pop}(\text{sort}(\text{chains}, \text{criteria}=$r$))$
   \FOR{paragraph $p$ {\bfseries in} $P$}
   \STATE $c' = \text{add\_step}(c, p)$
   \IF{$\text{gives\_answer}(c')$}
   \STATE $\text{complete.append}(c')$
   \ELSE
   \STATE $\text{chains.append}(c')$
   \ENDIF
   \ENDFOR
   \ENDWHILE
   \STATE \textbf{return} \text{complete}
\end{algorithmic}
\end{algorithm}


\end{document}